\pdfoutput=1

\documentclass[11pt]{article}

\usepackage[preprint]{acl}

\usepackage{times}
\usepackage{latexsym}

\usepackage[T1]{fontenc}

\usepackage[utf8]{inputenc}

\usepackage{microtype}
\usepackage{graphicx}
\usepackage{multirow}
\usepackage{booktabs}
\usepackage{amsmath}
\usepackage{inconsolata}

\title{Integrating Multi-scale Contextualized Information for Byte-based \\
Neural Machine Translation}

\author{Langlin Huang \textsuperscript{\rm 1,3}, Yang Feng \textsuperscript{\rm 1,2,3}\thanks{Corresponding author.} \\
  \textsuperscript{\rm 1} Key Laboratory of Intelligent Information Processing \\
  Institute of Computing Technology, Chinese Academy of Sciences \\
  \textsuperscript{\rm 2} Key Laboratory of AI Safety, Chinese Academy of Sciences \\
  \textsuperscript{\rm 3} University of Chinese Academy of Sciences \\
  \texttt{\href{mailto:h.langlin@wustl.edu}{h.langlin@wustl.edu}, \href{mailto:fengyang@ict.ac.cn}{fengyang@ict.ac.cn}} \\
  }

\begin{document}
\maketitle
\begin{abstract}

Subword tokenization is a common method for vocabulary building in Neural Machine Translation (NMT) models. However, increasingly complex tasks have revealed its disadvantages. First, a vocabulary cannot be modified once it is learned, making it hard to adapt to new words. Second, in multilingual translation, the imbalance in data volumes across different languages spreads to the vocabulary, exacerbating translations involving low-resource languages. 
While byte-based tokenization addresses these issues, byte-based models struggle with the low information density inherent in UTF-8 byte sequences. 
Previous works enhance token semantics through local contextualization but fail to select an appropriate contextualizing scope based on the input. Consequently, we propose the Multi-Scale Contextualization (MSC) method, which learns contextualized information of varying scales across different hidden state dimensions. It then leverages the attention module to dynamically integrate the multi-scale contextualized information. Experiments show that MSC significantly outperforms subword-based and other byte-based methods in both multilingual and out-of-domain scenarios. We have uploaded the code to github\footnote{\url{https://github.com/ictnlp/Multiscale-Contextualization}}.

\end{abstract}

\section{Introduction}

In neural machine translation (NMT) systems, subword tokenization has been the most common and effective method to mitigate the out-of-vocabulary (OOV) problem. However, both BPE~\cite{DBLP:conf/acl/SennrichHB16a} and SentencePiece~\cite{kudo-richardson-2018-sentencepiece} fix the word segmentation rule or vocabulary once they have learned them on the initial corpus, making it difficult to ensure adaptation to new corpora. This is worsened in out-of-domain scenarios.
Additionally, in multilingual scenarios with data imbalance, subword vocabularies tend to focus on high-resource languages, overlooking low-resource ones. This imbalance can cause an increase in OOV cases or over-segmentation of texts, harmful to translation model performance.

Byte-based method is able to solve these problems with few embedding parameters and has aroused extensive researches~\cite{DBLP:conf/aaai/WangCG20, shaham-levy-2021-neural, xue-etal-2022-byt5, DBLP:journals/corr/abs-2305-07185, DBLP:journals/corr/abs-2302-14220, sreedhar-etal-2023-local}. In byte-based models, text is converted into byte sequences according to UTF-8 encoding, with each byte as a token within the vocabulary. They generally use a vocabulary with a maximum size of 256 but can adapt to imbalanced scenarios like multilingual translation and out-of-domain adaptation. 

However, a feature of UTF-8 encoding hinders conventional Transformer model~\cite{DBLP:conf/nips/VaswaniSPUJGKP17} from adapting well to byte-based vocabulary: a single character may correspond to 1 to 4 UTF-8 bytes. The number is 1 for English characters, but Arabic and many Asian languages require multiple bytes to represent a single character. Therefore, sometimes a single byte does not have a determined meaning; it requires the integration of local information to encode its semantics. To address that, various methods have been proposed for integrating local contextual information. 
SU4MT~\cite{huang-etal-2023-enhancing} learns contextual information with an Attentive Semantic Fusion layer, but requires accurate segmentation.
MEGABYTE~\cite{DBLP:journals/corr/abs-2305-07185} segments a sentence into blocks of 4 and simply concatenates the tokens. Charformer~\cite{DBLP:conf/iclr/Tay0RGCB0B0M22} segments a sentence 4 times with block sizes ranging from 1 to 4 each, and employs mean-pooling to perform local integration. The weighted-sum of 4 results yields the final output. LOBEF~\cite{sreedhar-etal-2023-local} proposed Byte-\textit{n}CF, replacing mean-pooling in Charformer with Convolutional Neural Networks (CNNs) for better performance.

\begin{figure*}[h]
    \centering
    \includegraphics[width=\linewidth]{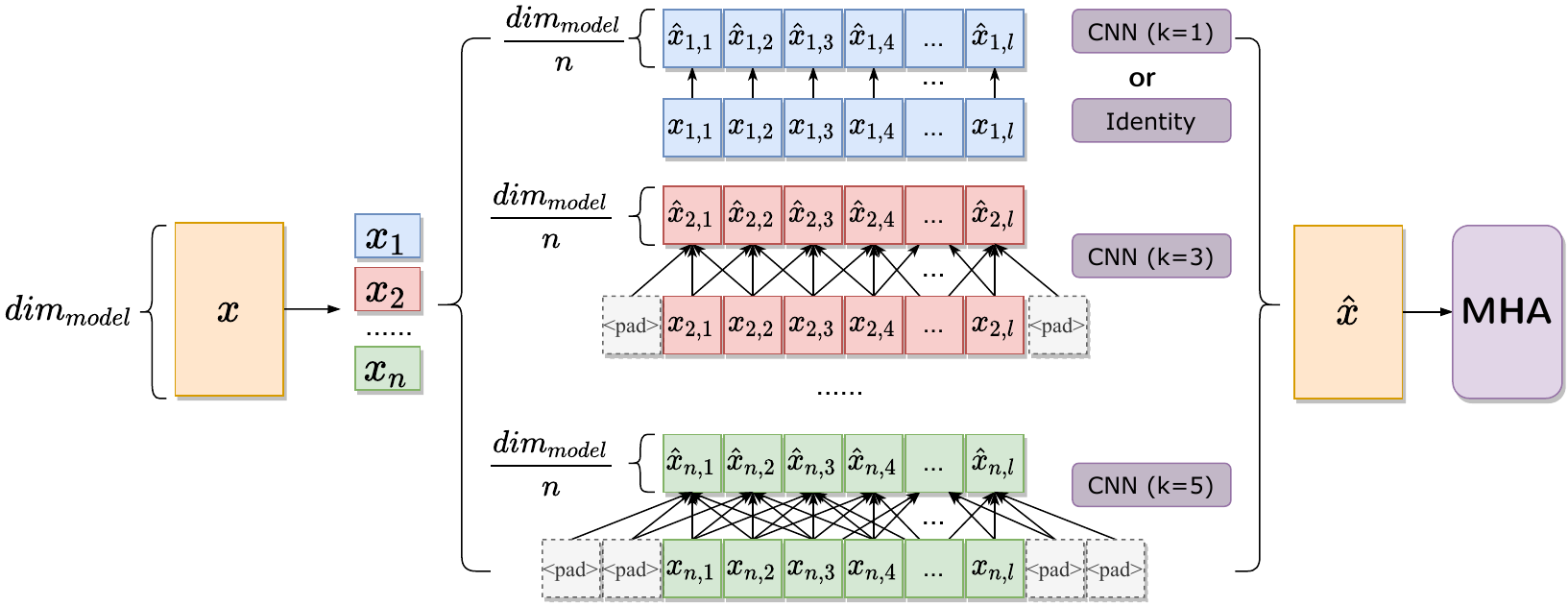}
    \caption{\textbf{Multi-Scale Contextualization} module: the input vector $x$, with hidden state dimension $dim_{model}$ and text length $l$, is divided into $n$ parts according to the hidden state dimensions, and then $n$ contextualizing functions with different scopes process these parts respectively. The output $\hat{x}$ now contains multi-scale information and acts as input to the Multi-Head Attention module.}
    \label{fig:model}
\end{figure*}

Though these methods learn and leverage contextual information in larger scales, they are limited by fixed block sizes and can not adjust the fusion weights according to the scripts of different languages. To remedy this, we propose the Multi-Scale Contextualization (MSC) method, which firstly learn contextual information of multiple scales in different hidden state dimensions, and secondly leverage the attention mechanism to fuse multi-scale information with dynamic weights. Our method better adapts to complex input scripts by allowing the model to adaptively fuse information of different granularities based on the varying content of the inputs.
Experimental results demonstrate that our MSC method exhibits superior adaptability across various languages and datasets.



\section{Method}
\label{sec:method}
In this section, we introduce the proposed Multi-Scale Contextualization (MSC) method. Byte-based models usually learn contextualized information implicitly. What MSC does is explicitly modeling the contextual information of multiple scales by grouping the hidden state dimensions and let different parts of the dimensions learn information of different scales.

Specifically, we insert a multi-scale contextualization module right before the Multi-Head Attention (MHA) module, as is depicted in Fig \ref{fig:model}.
The input vector $\textbf{x}$ is divided according to the hidden state dimension into $n$ parts $[\textbf{x}_1, \textbf{x}_2, ..., \textbf{x}_n]$. 

Then, $n$ contextualizing functions $g(\cdot)$ are applied to these parts respectively. A simple and effective structure for local contextualization is the 1-D convolution neural network (CNN). Therefore, we leverage CNNs with different kernel size $k$ to control the contextualization scope. Since different dimensions are contextualized with various granularities, our model can realize multi-scale contextualization. To preserve the original information, $g(\cdot)$ is also allowed to be a "Identity" function, which directly returns the input without any operations.
\begin{equation}
    g_i(\cdot, k)=\left\{
    \begin{aligned}
        & \mathrm{Identity}(\cdot) & , & ~k=0 \\
        & \mathrm{CNN}(\cdot,k) & , & ~k>0
    \end{aligned}
    \right.
    \label{eq:1}
\end{equation}

In equation \eqref{eq:1}, $g_i(\cdot, k)$ means the contextualization function for group $i$, and $k$ is the kernel size. Here, $k$=$0$ denotes the "Identity" function for simplification. 

Next, the contextualized vector parts $\hat{x}_i$ are calculated by $g_i(x_i, k)$. Finally, they are concatenated to form $\hat{x}$, which acts as the input of the MHA module. 

Preliminary experiments on CNN configurations guided the best structure for the whole model. First, padding on the left side deteriorates model performance, so the CNNs are set to pad on both sides, as shown in Figure \ref{fig:model}. Second, applying MSC to Transformer decoder layers causes a discrepancy between training and testing, when a token's right side tokens are not yet generated. As a result, MSC is only applied to encoder layers. 

\begin{table*}[h]
\centering
\scalebox{0.8}{
\begin{tabular}{c|ccc|ccc}
\hline
\toprule
\multirow{3}{*}{LID} & \multicolumn{3}{c|}{Subword}             & \multicolumn{3}{c}{Byte}                \\ \cline{2-7} 
                     & Learned        & mBART  & Aharoni$^*$        & Transformer    & Byte-\textit{n}CF   & MSC           \\
                     & (60.6M)          & (172.2M) & ($\sim$93M)              & (44.3M)          & (47.0M) & (45.0M)          \\ \hline
Az                   & 11.58 & 10.24  & 11.24          & $12.61_{(\pm0.42)}$ & $12.46_{(\pm0.27)}$ & $\textbf{13.24}_{(\pm0.25)}$          \\
Be                   & 18.83 & 15.41  & 18.28          & $21.41_{(\pm0.15)}$          & $21.51_{(\pm0.19)}$ & $\textbf{22.10}_{(\pm0.49)}$ \\
Gl                   & 26.81          & 28.20  & 28.63 & $31.08_{(\pm0.43)}$          & $31.44_{(\pm0.05)}$ & $\textbf{31.98}_{(\pm0.13)}$ \\
Sk                   & 24.93          & 24.76  & 26.78 & $27.64_{(\pm0.26)}$          & $27.65_{(\pm0.35)}$ & $\textbf{28.39}_{(\pm0.27)}$ \\
\textbf{AVG-LR}      & 20.54          & 19.65  & 21.23 & $23.19_{(\pm0.04)}$          & $23.26_{(\pm0.14)}$ & $\textbf{23.93}_{(\pm0.14)}$ \\ \hline
Ar                   & 23.35          & 22.57  & 25.93 & $25.60_{(\pm0.13)}$          & $25.89_{(\pm0.29)}$ & $\textbf{26.33}_{(\pm0.09)}$ \\
De                   & 26.33          & 27.78  & 28.87 & $30.14_{(\pm0.49)}$          & $30.56_{(\pm0.33)}$ & $\textbf{31.02}_{(\pm0.21)}$ \\
He                   & 27.09          & 26.59  & 30.19 & $30.38_{(\pm0.33)}$          & $30.55_{(\pm0.17)}$ & $\textbf{31.29}_{(\pm0.19)}$ \\
It                   & 28.45          & 30.36  & 32.42 & $32.97_{(\pm0.66)}$          & $33.34_{(\pm0.28)}$ & $\textbf{33.63}_{(\pm0.34)}$ \\
\textbf{AVG-HR}      & 26.31          & 26.83  & 29.35 & $29.77_{(\pm0.40)}$          & $30.08_{(\pm0.26)}$ & $\textbf{30.57}_{(\pm0.13)}$ \\ \hline
\textbf{AVG-58}      & 21.44          & 21.15  & -              & $23.63_{(\pm0.16)}$          & $23.70_{(\pm0.21)}$ & $\textbf{24.30}_{(\pm0.10)}$ \\ \hline
\end{tabular}
}
\caption{The experiment results on TED-59 dataset, measured by SacreBLEU. The table includes 4 low-resource (LR) and 4 high-resource (HR) languages selected by \citet{aharoni-etal-2019-massively}. The "$^*$" sign denotes the results are cited from \citet{aharoni-etal-2019-massively}. Byte-based models are experimented three times and the table shows the average scores and the standard deviation.}
\label{table:ted}
\end{table*}

It is worth noting that the kernel size $k$ is recommended to be an odd number or zero, otherwise, manual zero-padding is required to keep the output length the same as the input. Empirically, it is better to choose $k$ from \{0, 1, 3, 5, 7\}. 

\section{Experiments}

We experiment with two multilingual datasets and a domain-adaptation dataset to investigate the performances and properties of the MSC approach and other byte-based language models.

\subsection{Datasets}
\subsection*{Multilingual Many-to-One Translation}
We use a multilingual TED corpus of 59 languages~\cite{qi-etal-2018-pre}, TED-59, which includes both high and low-resource languages. All cases are English-centered. We collect the raw data from \citet{salesky-etal-2023-multilingual} and preprocess it with two subword-level vocabularies and a byte-level one. For subword-based baseline system, we leverage SentencePiece~\cite{kudo-richardson-2018-sentencepiece} and train a 32k vocabulary on the training set. We also incorporate the 250k mBART-50~\cite{liu-etal-2020-multilingual-denoising, DBLP:journals/corr/abs-2008-00401} vocabulary for full lexical coverage. \label{text:vocabulary}
For Byte-level systems, we preprocess data with a 256 vocabulary using scripts\footnote{\url{https://github.com/UriSha/EmbeddinglessNMT/blob/master/embeddingless_scripts/byte_preprocess.sh}} from \citet{shaham-levy-2021-neural}.

\subsection*{Multilingual English-Centric Translation}
We use the OPUS-7 corpus processed by \citet{gu-feng-2022-improving}, which is extracted from the OPUS-100 corpus~\cite{zhang-etal-2020-improving}. The OPUS-7 dataset contains a training corpus of 6 languages (Ar, De, Fr, Nl, Ru, Zh) and their English Translations, with 1M sentences of each language.

\begin{table*}[h]
\centering
\scalebox{0.66}{
\begin{tabular}{l|l|l|ccccccc|c}
\hline
Approach                     & Param.                 & Direction                & Ar    & De    & Fr    & Nl    & Ru    & Zh    & AVG  & AVG-all                   \\ \hline
Transformer- & \multirow{2}{*}{60.5M} & XX$\rightarrow$ En & $36.60$ & $34.02$ & $34.10$ & $30.28$ & $36.77$ & $38.82$ & $35.30$ & \multirow{2}{*}{$31.27$} \\
                        subword     &                        & En$\rightarrow$ XX & $21.61$ & $29.66$ & $31.84$ & $27.97$ & $30.70$ & $25.72$ & $27.92$ &                        \\ \hline \hline
Transformer- & \multirow{2}{*}{44.6M} & XX$\rightarrow$ En & $28.79_{(\pm0.30)}$ & $29.68_{(\pm0.11)}$ & $27.75_{(\pm0.21)}$ & $26.58_{(\pm0.04)}$ & $27.66_{(\pm0.30)}$ & $27.94_{(\pm0.25)}$ & $28.06_{(\pm0.14)}$ & \multirow{2}{*}{$25.19_{(\pm0.07)}$} \\
                        byte     &                        & En$\rightarrow$ XX & $13.51_{(\pm0.33)}$ & $26.45_{(\pm0.23)}$ & $25.09_{(\pm0.09)}$ & $24.00_{(\pm0.12)}$ & $18.28_{(\pm0.23)}$ & $26.59_{(\pm0.20)}$ & $22.32_{(\pm0.03)}$ &                        \\ \hline
\multirow{2}{*}{Byte-nCF}    & \multirow{2}{*}{47.0M} & XX$\rightarrow$ En & $\textbf{31.19}_{(\pm0.31)}$ & $\textbf{30.86}_{(\pm0.33)}$ & $29.28_{(\pm0.28)}$ & $27.76_{(\pm0.20)}$ & $29.17_{(\pm0.23)}$ & $29.65_{(\pm0.17)}$ & $29.65_{(\pm0.22)}$ & \multirow{2}{*}{$26.59_{(\pm0.19)}$} \\
                             &                        & En$\rightarrow$ XX & $14.58_{(\pm0.14)}$ & $27.58_{(\pm0.21)}$ & $26.20_{(\pm0.14)}$ & $25.05_{(\pm0.22)}$ & $\textbf{19.60}_{(\pm0.51)}$ & $28.17_{(\pm0.20)}$ & $23.53_{(\pm0.17)}$ &                        \\ \hline
\multirow{2}{*}{MSC}         & \multirow{2}{*}{44.8M} & XX$\rightarrow$ En & $31.16_{(\pm0.11)}$ & $\textbf{30.86}_{(\pm0.22)}$ & $\textbf{29.31}_{(\pm0.06)}$ & $\textbf{28.10}_{(\pm0.13)}$ & $\textbf{29.49}_{(\pm0.15)}$ & $\textbf{29.75}_{(\pm0.18)}$ & $\textbf{29.78}_{(\pm0.05)}$ & \multirow{2}{*}{$\textbf{26.79}_{(\pm0.06)}$} \\
                             &                        & En$\rightarrow$ XX & $\textbf{14.86}_{(\pm0.12)}$ & $\textbf{27.89}_{(\pm0.47)}$ & $\textbf{26.62}_{(\pm0.15)}$ & $\textbf{25.39}_{(\pm0.23)}$ & $19.57_{(\pm0.18)}$ & $\textbf{28.43}_{(\pm0.34)}$ & $\textbf{23.80}_{(\pm0.16)}$ &                        \\ \hline
\end{tabular}
}
\caption{The experiment results on OPUS-7 dataset, measured by SacreBLEU. All approaches are trained on the 12 directions together. Byte-based models are experimented three times and the table shows the average scores and the standard deviation. }
\label{table:opus7}
\end{table*}

\subsection*{Zero-shot Cross-domain Adaptation}
Besides multilingual scenarios, we also experiment with the zero-shot cross-domain adaptation ability of byte-based translation models with the WMT19 German$\rightarrow$English (De$\rightarrow$En) dataset. We train all models on the News domain and evaluate on test data from three domains used in \citet{sreedhar-etal-2023-local} and \citet{aharoni-goldberg-2020-unsupervised}
, which are Koran, IT, and Medical. We use the data preprocessed and provided by \citet{sreedhar-etal-2023-local}.

\subsection{Models}
We compare the proposed MSC approach mainly with other byte-based machine translation models. 
\begin{itemize}
    \item \textbf{Transformer}~\cite{DBLP:conf/nips/VaswaniSPUJGKP17}: The standard Transformer model without adaptations to byte sequences.
    \item \textbf{Byte-\textit{n}CF}~\cite{sreedhar-etal-2023-local}: A strong byte-based model performing well under low-resource settings. The structure hyper-parameters are of default setting\footnote{\url{https://github.com/makeshn/LOBEF_Byte_NMT/blob/main/embeddingless_scripts/train_byte_ncf.sh}}.
    \item \textbf{MSC}: We set $n$=$8$ in our experiments. The selection of $k$ is discussed in Appendix \ref{appendix:k-selection}.
\end{itemize}
We also compare with subword-based methods.
\begin{itemize}
    \item \textbf{Learned}: The standard Transformer model with a learned vocabulary.
    \item \textbf{mBART}: The standard Transformer model using the vocabulary of mBART~\cite{liu-etal-2020-multilingual-denoising}. We do not use the pretrained checkpoint of mBART for fairness.
    \item \textbf{Aharoni}: The strongest baseline of subword-based models in this parameter scale.
\end{itemize}

The other settings are discussed in Appendix \ref{appendix:model-setting}.

\section{Results and Analyses}
\subsection{Multilingual Many-to-One Translation}
Table \ref{table:ted} shows the results on TED-59 datasets. All byte-based methods are experimented three times to enhance the reliability of results. We report the average and standard deviation of the results. \citet{aharoni-etal-2019-massively} have selected four low-resource (LR) and four high-resource (HR) languages to show models' performance on different training data scales, and we report results in the same way.

The average SacreBLEU scores of 58 translation directions (AVG-58) demonstrate that byte-based models are superior to subword-based models in massively multilingual scenario, despite of lower parameter usage. 

Compared with other byte-based approaches, MSC performs better in almost all languages. 
While Byte-\textit{n}CF learns a fixed set of combination weights of multi-scale contextual information for all languages,
MSC adaptively leverages contextual information of different granularities at inference stage.
For example, a single byte can represent a character or even a word in German, Italian, etc., so MSC leverages contextual information from its nearer neighborhood; a single byte may not be sufficient to form even a character, so MSC inclines to focus on contextual information of larger scales. We demonstrate this explanation later with an experiment in \ref{subsec:contextualization-scales}.

\subsection{Multilingual English-Centric Translation}
Table \ref{table:opus7} shows the results on OPUS-7 dataset, which contains only seven high-resource languages. In this scenario, the subword-based model largely surpasses byte-based models. However, the performance gap is smaller when measured by COMET~\cite{rei-etal-2022-comet}, a more reliable model-based metric, as reported in Appendix~\ref{appendix:results-in-comet}.

Among these byte-based models, MSC generally performs better than the others. To verify such improvements are not from randomness, we repeat them for three times and report the average and standard deviation of the results.

\begin{table*}[h]
\centering
\scalebox{0.8}{
\begin{tabular}{l|ccc|cccc}
\hline
\multirow{3}{*}{Domain} & Transformer- & Transformer- & \multirow{2}{*}{Byte-nCF$^*$} & Transformer- & Transformer- & \multirow{2}{*}{Byte-nCF} & \multirow{2}{*}{MSC} \\
                        & subword$^*$   & byte$^*$      &                           & subword   & byte      &                           &                      \\  
                        & (68.7M)     & (44.3M)     & (46.7M)                   & (64.6M)     & (44.2M)     & (46.8M)                   & (44.7M)              \\ \hline
News                    & 17.6        & 21.2        & 21.3                      & 21.06       & 21.58       & 21.81                     & \textbf{21.86}       \\ \hline
Koran                   & 1.8         & 6.6         & \textbf{7.4}              & 1.46        & 6.74        & 6.58                      & 6.83                 \\
IT                      & 3.9         & 10.4        & 11.6                      & 2.73        & 10.89       & 11.33                     & \textbf{12.49}$^\Uparrow$       \\
Medical                 & 4.1         & 13.6        & 15.3                      & 2.79        & 17.19       & 17.41                     & \textbf{20.01}$^\Uparrow$       \\ \hline
AVG                    & 3.27        & 10.20       & 11.43                     & 2.33        & 11.61       & 11.77                     & \textbf{13.11}       \\ \hline
\end{tabular}
}
\caption{The experiment results on WMT19 De$\rightarrow$En domain adaptation dataset. The "$^*$" sign denotes the results are cited from \citet{sreedhar-etal-2023-local}. The average results of Koran, IT, and Medical domains indicate byte-based models perform better than subword-based models when test sets contain many rare words or even unknown words. The "$\Uparrow$" sign denotes MSC prominently outperforms the second best method, with "$p<0.001$".}
\label{table:domain}
\end{table*}

\subsection{Zero-Shot Cross-Domain Adaptation}
Table \ref{table:domain} shows the results on in-domain and zero-shot out-of-domain test datasets. 
For subword-based models, using the dictionary trained on the News domain dataset to preprocess test sets from other domains results in a significant number of \textit{<unk>} words. This leads to the model struggling to comprehend the input sentences and perform translations. However, byte-based models can eliminate the Out-Of-Vocabulary (OOV) issue, thereby achieving better performance in zero-shot translation scenarios. The results also demonstrate that our proposed MSC method has a significant advantage in zero-shot cross-domain adaptation.

\subsection{Contextualization Scales}
\label{subsec:contextualization-scales}
In section \ref{sec:method}, we have introduced the hyper-parameter $k$ which controls the contextualization scale of our approach. Here, we experiment on TED-59 dataset to show how this modeling scale affect translation.

According to the Unicode rule, we group languages by the number of bytes they require to form a character, which are named "Byte-1", "Byte-2", and "Byte-3". Then, we select three languages for each group to represent that group, as listed below. 
\begin{itemize}
    \item Byte-1: French (Fr), German (De), Dutch (Nl)
    \item Byte-2: Russian (Ru), Thai (Th), Arabic (Ar)
    \item Byte-3: Chinese (Zh), Japanese (Ja), Korean (Ko)
\end{itemize}

For the selection of $k$ series, which reflect the contextualization scales applied in a model, we experiment the \textbf{small} scales "0,0,1,1,3,3,5,5", the \textbf{large} scales "0,0,1,1,5,5,7,7", and the \textbf{balanced} scales "0,0,1,1,3,5,5,7". These models can leverage information of granularities of "1,3,5", "1,5,7", and "1,3,5,7" respectively\footnote{We have also experimented the granularities "9", "11" and "13", but they are harmful to translation quality. It demonstrates the model is unable to process information from too many tokens well.}.

\begin{table}[h]
\centering
\begin{tabular}{l|c|c||c}
\hline
 & small &  large       & balanced         \\ \hline
Byte-1    & 30.190*  &  29.941 & \textbf{30.219}           \\ \hline
Byte-2    & 21.251   &   \textbf{21.601*}    & 21.545 \\ \hline
Byte-3    & 10.302   & 10.686* &    \textbf{10.712}       \\ \hline
\end{tabular}
\caption{The selection of hyper-parameter $k$ series affects model performance of different language groups.}
\label{table:msa-analyse-k}
\end{table}

The results are exhibited in Table \ref{table:msa-analyse-k}. First, the model of balanced scales performs the best averagely, because it is provided with contextualized information of more scales and has more options to choose. Second, if we ignore the balanced one and compare the other models, the performance is related to the language groups. The "$^*$" sign indicates the better performance between two models. For "Byte-1" group, a small scale is sufficient to model certain semantics, so the smaller scaled model performs better. For "Byte-3" groups, it requires larger contextual scales to form certain meanings, so the larger scaled model performs better.

These discoveries shade light on the selection of $k$ series, which is the $k$ series should be compatible with the language of input text.


\section{Conclusions}
In this paper, we make two primary contributions. Firstly, we show when byte-based models outperforms subword-based models and when they don't. In massively multilingual translation scenarios which involves a wide range of languages, byte-based models exhibit clear superiority, particularly for low-resource languages. In limited language numbers with sufficient training data, byte-based models lags behind subword-based models. Secondly, we introduce a Multi-Scale Contextualization (MSC) method which enhances the adaptability of byte-based models to diverse inputs. By dynamically integrating multi-scale contextualized information, MSC outperforms other byte-based models in generally all languages.

\section*{Limitations}
While our approach can adaptively integrate multi-scale contextualized information, all contextualizing scopes $k$ are predetermined. We will explore fully adaptive methods for multi-scale information extraction and integration for future work.

\section*{Acknowledgements}

We thank all the anonymous reviewers for their
insightful and valuable comments. This paper is supported by National Natural Science Foundation of China (Grant No.62376260).
\bibliography{custom}

\newpage
\onecolumn
\appendix


\section{Selection of $k$ Series}
\label{appendix:k-selection}
The selection of $k$ varied across different languages. 
A critical determinant is the number of UTF-8 bytes required to represent a character. For languages that use the Latin alphabet, where a single byte can represent a character, a smaller $k$ suffices; conversely, for languages where a character corresponds to multiple bytes, the information density of a single byte is lower, necessitating a larger 
$k$. Additionally, we find the "PassThrough$(\cdot)$" function indispensable as it preserves the original information in the hidden states.

Empirically, the $k$ series is "0,0,1,1,3,3,5,5" for De$\rightarrow$En domain adaptation dataset, "0,0,1,1,3,5,5,7" for OPUS-7 dataset, and "0,0,3,3,5,5,7,7" for TED-59 dataset.

\section{Detailed Model Settings}
\label{appendix:model-setting}
For a fair comparison, all model implementations are based on the Fairseq~\cite{ott-etal-2019-fairseq} codebase. In our experiments, all models contain 6 transformer encoder layers and 6 transformer decoder layers with 8 attention heads. The model dimension is 512 for word embedding and 2048 for feed-forward layers. The word embeddings of the encoder and decoder are shared for all models. For our method, the MSC module is applied to the first encoder layer. For Byte-\textit{n}CF, we use the default model structure, but apply the shared-embedding setting, which on one hand prove to be better than its default settings, and on the other hand align with other experiments.

For model training, we train all models using 8 GPUs with batch size 8192 for each. We use adam optimizer~\cite{DBLP:journals/corr/KingmaB14} with $\beta$=$(0.9, 0.98)$ and 4k warm-up steps. The peak learning rate is $5e$-$4$ in multilingual tasks and $7e$-$4$ in De$\rightarrow$En cross-domain adaptation task. Besides, we apply dropout 0.1 and label smoothing 0.1 for all models. We apply an early stop of 10, and average the last 5 checkpoints for evaluation. All models are evaluated using the SacreBLEU score.

\section{Experiment Results Measured by COMET}
\label{appendix:results-in-comet}

\begin{table*}[h]
\centering
\begin{tabular}{c|ccc}
\hline
\multirow{2}{*}{LID} & Transformer  & Byte-nCF     & MSC          \\
                     & (44.3M)      & (47.0M)      & (45.0M)      \\ \hline
az                   & $68.34_{(\pm0.72)}$ & $67.47_{(\pm0.49)}$ & $\textbf{69.27}_{(\pm0.42)}$ \\
be                   & $70.52_{(\pm0.80)}$ & $70.06_{(\pm0.27)}$ & $\textbf{71.33}_{(\pm0.47)}$ \\
gl                   & $78.00_{(\pm0.17)}$ & $78.04_{(\pm0.20)}$ & $\textbf{78.61}_{(\pm0.13)}$ \\
sk                   & $76.49_{(\pm0.18)}$ & $76.39_{(\pm0.46)}$ & $\textbf{77.39}_{(\pm0.38)}$ \\
\textbf{AVG-LR}               & $73.34_{(\pm0.38)}$ & $72.99_{(\pm0.35)}$ & $\textbf{74.15}_{(\pm0.33)}$ \\ \hline
ar                   & $74.13_{(\pm0.51)}$ & $74.00_{(\pm0.20)}$ & $\textbf{74.74}_{(\pm0.59)}$ \\
de                   & $76.88_{(\pm0.24)}$ & $76.93_{(\pm0.36)}$ & $\textbf{77.77}_{(\pm0.27)}$ \\
he                   & $75.82_{(\pm0.49)}$ & $75.65_{(\pm0.27)}$ & $\textbf{76.63}_{(\pm0.63)}$ \\
it                   & $78.43_{(\pm0.25)}$ & $78.67_{(\pm0.20)}$ & $\textbf{79.12}_{(\pm0.16)}$ \\
\textbf{AVG-HR}               & $76.31_{(\pm0.23)}$ & $76.32_{(\pm0.25)}$ & $\textbf{77.06}_{(\pm0.41)}$ \\ \hline
\textbf{AVG-58}               & $73.00_{(\pm0.37)}$ & $74.65_{(\pm0.30)}$ & $\textbf{75.61}_{(\pm0.37)}$ \\ \hline
\end{tabular}
\caption{The experiment results on TED-59 dataset, measured by COMET.}
\label{table:ted-comet}
\end{table*}

\begin{table*}[h]
\centering
\scalebox{0.66}{
\begin{tabular}{l|l|l|ccccccc|c}
\hline
Approach                     & Param.                 & Direction                & Ar    & De    & Fr    & Nl    & Ru    & Zh    & AVG  & AVG-all                   \\ \hline
Transformer- & \multirow{2}{*}{60.5M} & XX$\rightarrow$ En & $79.03$ & $80.10$ & $79.27$ & $78.01$ & $78.14$ & $79.15$ & $78.89$ & \multirow{2}{*}{$78.65$} \\
                        subword     &                        & En$\rightarrow$ XX & $78.50$ & $78.41$ & $76.94$ & $78.05$ & $79.09$ & $79.72$ & $78.45$ &                        \\ \hline \hline
Transformer- & \multirow{2}{*}{44.6M} & XX$\rightarrow$ En & $77.31_{(\pm0.16)}$ & $77.69_{(\pm0.04)}$ & $77.16_{(\pm0.06)}$ & $75.98_{(\pm0.04)}$ & $75.64_{(\pm0.12)}$ & $76.66_{(\pm0.15)}$ & $76.74_{(\pm0.06)}$ & \multirow{2}{*}{$76.19_{(\pm0.04)}$} \\
                        byte     &                        & En$\rightarrow$ XX & $75.00_{(\pm0.10)}$ & $74.37_{(\pm0.02)}$ & $72.51_{(\pm0.05)}$ & $74.11_{(\pm0.06)}$ & $71.38_{(\pm0.05)}$ & $86.46_{(\pm0.08)}$ & $75.64_{(\pm0.02)}$ &                        \\ \hline
\multirow{2}{*}{Byte-nCF}    & \multirow{2}{*}{47.0M} & XX$\rightarrow$ En & $ 78.68_{(\pm0.25)}$ & $78.62_{(\pm0.10)}$ & $78.24_{(\pm0.20)}$ & $77.18_{(\pm0.31)}$ & $76.77_{(\pm0.17)}$ & $77.88_{(\pm0.42)}$ & $77.90_{(\pm0.23)}$ & \multirow{2}{*}{$77.55_{(\pm0.26)}$} \\
                             &                        & En$\rightarrow$ XX & $76.35_{(\pm0.24)}$ & $76.08_{(\pm0.18)}$ & $73.85_{(\pm0.37)}$ & $75.61_{(\pm0.34)}$ & $73.73_{(\pm0.40)}$ & $87.63_{(\pm0.27)}$ & $77.21_{(\pm0.29)}$ &                        \\ \hline
\multirow{2}{*}{MSC}         & \multirow{2}{*}{44.8M} & XX$\rightarrow$ En & $\textbf{78.90}_{(\pm0.06)}$ & $\textbf{78.84}_{(\pm0.07)}$ & $\textbf{78.32}_{(\pm0.05)}$ & $\textbf{77.37}_{(\pm0.07)}$ & $\textbf{76.92}_{(\pm0.13)}$ & $\textbf{78.09}_{(\pm0.08)}$ & $\textbf{78.08}_{(\pm0.06)}$ & \multirow{2}{*}{$\textbf{77.85}_{(\pm0.03)}$} \\
                             &                        & En$\rightarrow$ XX & $\textbf{76.77}_{(\pm0.06)}$ & $\textbf{76.36}_{(\pm0.10)}$ & $\textbf{74.32}_{(\pm0.15)}$ & $\textbf{76.08}_{(\pm0.03)}$ & $\textbf{74.28}_{(\pm0.07)}$ & $\textbf{87.92}_{(\pm0.17)}$ & $\textbf{77.62}_{(\pm0.02)}$ &                        \\ \hline
\end{tabular}
}
\caption{The experiment results on OPUS-7 dataset, measured by COMET.}
\label{table:opus7-comet}
\end{table*}

\end{document}